# M2M 통신을 사용한 트래픽 모니터링

시우 쿠마, 조용옥, 함은식*, 신준우*, 이성로

목포대학교, *정보통신산업진흥원

shiu748@gmail.com, hotlove555kr@naver.com, srlee@mokpo.ac.kr

# Traffic Monitoring Using M2M Communication

Shiu Kumar, Yong Ok Jo, Eun Sik Ham*, Jun Woo Shin* and Seong Ro Lee

Mokpo National Univ., *NIPA

## Abstract

This paper presents an intelligent traffic monitoring system using wireless vision sensor network that captures and processes the real-time video image to obtain the traffic flow rate and vehicle speeds along different urban roadways. This system will display the traffic states on the front roadways that can guide the drivers to select the right way and avoid potential traffic congestions. On the other hand, it will also monitor the vehicle speeds and store the vehicle details, for those breaking the roadway speed limits, in its database. The real-time traffic data is processed by the Personal Computer (PC) at the sub roadway station and the traffic flow rate data is transmitted to the main roadway station Arduino 3G via email, where the data is extracted and traffic flow rate displayed.

## I. Introduction

Increasing congestion level in public road networks is a growing problem in many countries. Traffic problems have stagnated along with the economy; an annual study [1] suggests that too little improvement is being made toward ensuring that the nation's transportation system will be able to sustain with job growth when the economy does return.

Inductive loop [2] and magnetic sensors [3] have been conventionally used for traffic management and control applications, however, destruction of roadbed and limited spatial sensing are two major limitations. Other methods such as Traffic-Dot [4], which consist of wireless sensor network and access point, have also been tested. However, due to the significant advances in the field of computer vision recently, traffic control and monitoring using vision sensors has drawn increasing attention. Many commercial and research systems [5] and [6] use video processing, aiming to solve specific problems in road traffic monitoring. Traffic congestion results in direct economic loss, and also exacerbates the air pollution of the urban cities because vehicles emit more carbon dioxide and other toxic materials when they are at a lower speed. This paper intends to alleviate this situation by piloting the vehicles to the unblocked roads using real time traffic state monitoring.

## II. The framework for traffic monitoring using wireless vision sensors.

The traffic monitoring systems is composed of two stations; sub roadways and main roadway. The sub roadway station consists of a video camera and a PC connected to the internet (Arduino 3G module can also be used as a modem for internet connectivity), while the main roadway comprises of a 3G Arduino mega 2560, and a display. The real-time video of the traffic flow is captured and processed by the PC to obtain the vehicle speed and the traffic flow rate. The traffic flow rate data is then transmitted to the Arduino 3G at the main roadway station via an email. The data received is then manipulated and traffic flow rate is displayed. The proposed system architecture is shown in figure 1.

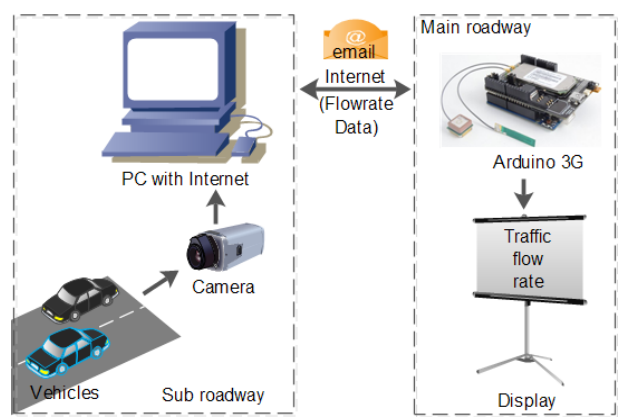

Fig. 1: Traffic monitoring system architecture

## III. Video processing and data extraction

The real-time video is captured using a high resolution camera, which is connected to the PC. The video processing is done using OpenCV library alongside Visual Studio 2010. The moving vehicles are detected by the use of the absolute difference function in OpenCV. By calculating the absolute difference between the current frame and the frame before, only the movement is shown. The frames are converted to gray scale and passed through a threshold filter, which makes it possible to track movement in the specific area of the image by checking for the white pixels. The processed image is further passed through a Gaussian filter to smooth the image, which makes it easier for the detection of moving vehicles in the next step.

The moving vehicles are thus detected and the time taken, $t_{AB}$, for each vehicle to travel a known distance, L, from point A to point B as shown in figure 2 is recorded. The centre point of the vehicles is used for all measurements. This time is used to determine the average vehicle speed; however the vehicle speed is also estimated as the vehicle moves along the path from point A to point B. If the vehicle speed is over the roadway speed limit specified, the vehicle image and speed data together with date and time are recorded in a database, where the image is further processed to obtain the vehicle registration. The distance L is determined by experimentation using the camera so that vehicles are accurately detected, and to a great extent is dependent on the type of camera used.

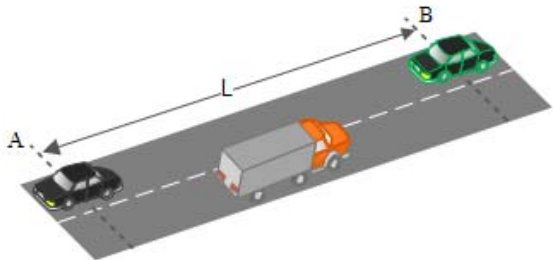

Fig. 2: Roadway layout as seen by the Camera

As the vehicles passes point B, the vehicle count, increments taking note of the number of vehicles passing by in a specified time interval, t. The time interval t depends on the system designer and can be changed for different sub roadways. Then the average traffic flow rate is determined using equation 1 and the data is emailed to the main roadway Arduino 3G using visual C++ platform. The vehicle count is then reset and starts count for the next interval.

$$Traffic\ flow\ rate = \frac{Vehicle\ Count}{t}\quad (vehicles/\min)\qquad (1)$$

The Arduino 3G at the main roadway station sits idle as long as no data is available at the input port. Once the data is available, that is it receives the email from the sub roadway station, the traffic flow rate data is extracted from the email and appropriate control signals are generated to display the traffic flow rate.

The traffic flow-rate of all the sub roadways is thus displayed on the main roadway, providing information to the vehicle drivers to select the right route to travel thus avoiding congestion. The displayed data is updated as soon as the Arduino 3G receives the next traffic flow rate data via the email.

The system is also capable of being interrupted, where the traffic personal can send an email to the Arduino 3G (this is the email address configured by the program algorithm), thus the email message will be displayed instead of the traffic flow rate. However, the email will also carry a security code for this interruption for security purposes. During this time, the Arduino 3G will automatically send an email informing the sub roadway station to pause traffic monitoring thus avoiding unnecessary processing. The same applies for resetting the traffic monitoring process.

## IV.    Conclusion

A real time monitoring system designed to monitor the traffic flow rate and vehicle speeds improves the situation of traffic congestion and air pollution on busy urban roadways. The system is also equipped with a security interrupt, which can be used to convey information, such as road closure due to accident or repair works, about a particular sub roadway to the vehicle drivers. This leads to vehicles having to go through minimum congestion travelling at higher speeds and thus also accounts for less fuel consumption.

## Acknowledgment

" 이 논문은 2011년도 정부(교육과학기술부)의 재원으로 한국연구재단의 지원을 받아 수행된 연구임(No.2011-0029321).", "본 연구는 지식경제부 및 정보통신산업진흥원의 대학 IT 연구센터 지원사업의 연구결과로 수행되었음"(NIPA-2012-H0301-12-2005)